# Automated Lane Change Behavior Prediction and Environmental Perception Based on SLAM Technology

Han Lei[1]*,Baoming Wang[1.2],Zuwei Shui[2],Peiyuan Yang[3],Penghao Liang[4]

1*Computer Science Engineering,Santa Clara University,Santa Clara, USA

1.2 Electrical and Computer Engineering,University of Illinois Urbana-Champaign,Urbana, IL, USA

2 Information Studies,Trine University, Phoenix, USA

3Electrical and Computer Engineering,University of Illinois at Urbana-Champaign,Champaign, Illinois, USA

4 Information Systems,Northeastern University,Boston, MA, USA

**Corresponding author:**[Han Lei,E-mail:hannahleigh19970807@gmail.com]

**Abstract.**

In addition to environmental perception sensors such as cameras, radars, etc. in the automatic driving system, the external environment of the vehicle is perceived, in fact, there is also a perception sensor that has been silently dedicated in the system, that is, the positioning module. This paper explores the application of SLAM (Simultaneous Localization and Mapping) technology in the context of automatic lane change behavior prediction and environment perception for autonomous vehicles. It discusses the limitations of traditional positioning methods, introduces SLAM technology, and compares LIDAR SLAM with visual SLAM. Real-world examples from companies like Tesla, Waymo, and Mobileye showcase the integration of AI-driven technologies, sensor fusion, and SLAM in autonomous driving systems. The paper then delves into the specifics of SLAM algorithms, sensor technologies, and the importance of automatic lane changes in driving safety and efficiency. It highlights Tesla's recent update to its Autopilot system, which incorporates automatic lane change functionality using SLAM technology. The paper concludes by emphasizing the crucial role of SLAM in enabling accurate environment perception, positioning, and decision-making for autonomous vehicles, ultimately enhancing safety and driving experience.

## 1. Introduction

In the development of autonomous vehicles, accurate positioning and mapping solutions are critical, especially in situations involving automatic lane changes and environmental awareness. In order to achieve this goal, SLAM technology has become an important solution. Traditionally, LIDAR and camera sensors have been widely used for positioning and sensing tasks in vehicles. However, the development of liDAR SLAM methods does not seem to have changed much over time. In contrast, visual SLAM technology offers many advantages, including low cost, ease of installation, and powerful scene recognition capabilities. Therefore, in recent years, people have begun to explore the application of cameras to autonomous vehicles to realize the perception and positioning of complex tasks such as automatic lane changes. In this context, automatic lane change behavior prediction and environment perception based on SLAM technology are reviewed in this paper. Through the detailed discussion of visual SLAM, we will focus on its application in automatic lane change behavior prediction and environment perception. By studying and discussing these technologies, we can better understand their potential in the field of autonomous driving and point the way for future development.

## 2. Related work

*2.1. Artificial intelligence and autonomous driving*

With the development of robotics and artificial intelligence (AI) technology, autonomous vehicles (AVs) have become a hot topic in both industry and academia. Safely navigating requires creating an accurate representation of the surrounding environment and estimating the vehicle's state within it, particularly its positioning. Traditional methods, relying on GPS or real-time kinematic (RTK) positioning systems, have limitations due to signal reflection, time errors, and atmospheric conditions. GPS, for instance, suffers from measurement errors up to a dozen meters, unsuitable for precise navigation, especially in complex urban or tunnel scenarios. Although RTK can mitigate these errors by calibrating with base stations, it involves expensive infrastructure.

For AVs on highways, precise vehicle positioning is paramount. This entails integrating data from multiple sensors and fusing it with map information. Highway positioning involves three main modules: inferring the current road, estimating the vehicle's lane position, and ensuring safe driving maneuvers like overtaking. To address these challenges, we present a taxonomy of localization methods tailored for highway scenarios, examining each component of the localization process and evaluating the latest methods.

Simultaneous Localization and Mapping (SLAM) stands out as a promising solution for AV positioning and navigation. SLAM can estimate vehicle attitude in real time while mapping the surrounding environment. Depending on sensor type, SLAM methods are categorized into LIDAR SLAM and Visual SLAM. LIDAR SLAM, being established earlier, is relatively mature for autopilot applications. Unlike cameras, LIDAR sensors are less sensitive to light changes and perform well in low-light conditions, providing a broader field of view and 3D mapping capabilities. However, the high cost and long development cycles hinder widespread adoption of LIDAR sensors.

Conversely, Visual SLAM offers advantages such as informativeness, ease of installation, and cost-effectiveness. Modern visual SLAM systems can run on micro PCs and embedded devices, even on smartphones. This accessibility contributes to its popularity and integration into various autonomous driving systems.

Now, let's delve into some real-world examples showcasing the synergy between AI and autonomous driving:

1. Tesla Autopilot: Tesla's Autopilot system employs AI and computer vision algorithms to enable features like adaptive cruise control, lane-centering, and automatic lane changes. By integrating data from cameras, radar, and ultrasonic sensors, it provides advanced driver assistance capabilities.

2. Waymo: Waymo's self-driving technology utilizes a combination of LIDAR, radar, and cameras for perception and mapping. Its AI algorithms interpret sensor data in real-time to make driving decisions, enabling fully autonomous driving in certain conditions.

3. Mobileye: Mobileye, an Intel company, develops advanced driver assistance systems (ADAS) and autonomous driving technology. Its EyeQ system-on-chip (SoC) integrates camera inputs with AI algorithms to provide features like lane-keeping assist and pedestrian detection.

These examples highlight how AI-driven technologies like SLAM, coupled with sensor fusion, are driving the evolution of autonomous driving systems, making them safer and more efficient on the road.

*2.2. SLAM*

SLAM (Simultaneous Localization and Mapping), also known as [1]CML (Concurrent Mapping and Localization), is a simultaneous mapping and mapping process. The problem can be described as: Put a robot in an unknown location in an unknown environment, is there a way for the robot to gradually draw a complete map of the environment, and at the same time decide which direction the robot should go. For example, a robot vacuum cleaner is a typical SLAM problem, and a consistent map means traveling to every accessible corner of the room without obstacles.

SLAM was first proposed by Smith, Self, and Cheeseman in 1988. Because of its important theoretical and application value, it is considered by many scholars to be the key to realize truly fully autonomous mobile robots.

When human beings come to a strange environment, in order to quickly familiarize themselves with the environment and complete their tasks (such as finding a restaurant, finding a hotel), they should do the following things in turn:

a. Observe the surrounding landmarks such as buildings, trees, flower beds, etc., with your eyes, and remember their features (feature extraction)

b. Reconstruct the feature landmark in the 3D map in your mind according to the information obtained by the eyes (3D reconstruction)

c. Constantly acquire new feature landmarks while walking, and adjust the map model in your mind (bundle adjustment or EKF)

d. Determine your trajectory according to the characteristic landmarks you obtained from walking some time ago.

e. When you inadvertently walk a long way, match the past landmarks in your mind to see if you have returned to the original path (loop-closure detection). The actual step is optional.

**Here are some specifics regarding the SLAM algorithm utilized in autonomous driving:**

1. Core function: The core function of SLAM algorithm is that it can locate the position of the vehicle itself in real time in the process of automatic driving, and build a map of the surrounding environment. This process takes place simultaneously, that is, updating the map of the environment while positioning itself, and vice versa.

2. Feature recognition: In automatic driving applications, SLAM usually needs to identify low-level feature points in images, and obtain the scene structure and pose information of the camera itself, namely position and attitude, by calculating the position changes of these feature points on different image frames.

3. Internal and external parameters: For SLAM systems using cameras as sensors, it is very important to understand the internal parameters of the camera (such as film size, resolution, optical lens coefficient, etc.) and external parameters (that is, the position and orientation Angle of the camera in the world coordinate system). These parameters help to accurately estimate the motion of the camera and correctly project three-dimensional space points onto the two-dimensional image plane.

4. Advanced features and deep learning: [2]In addition to feature-point based SLAM, modern autonomous driving systems also use deep learning models (such as convolutional neural networks CNN and Transformer) to identify advanced feature information such as vehicles, roads, pedestrians and obstacles, which helps to improve the accuracy and robustness of recognition.

5. Usability and accuracy: An effective SLAM algorithm needs to be able to accurately and timely provide the accuracy required for positioning, which is crucial for the safety of autonomous driving. In addition, a good SLAM system should be able to utilize existing map resources or integrate global map information to optimize location and navigation performance.

At present, the sensors used in SLAM are mainly divided into LiDAR and vision sensors:

Laser SLAM uses single-line or multi-line liDAR, which is generally used in indoor robots and unmanned fields. The emergence and popularity of Lidar makes the measurement faster and more accurate, and the information is richer. [3]The object information collected by LiDAR presents a series of scattered points with accurate Angle and distance information, known as point clouds. Usually, the laser SLAM system calculates the relative moving distance and attitude change of the laser radar by matching and comparing the two point clouds at different times, and then completes the positioning of the robot itself.

*2.3. Visual SLAM*

Visual SLAM can obtain massive and redundant texture information from the environment, and has superior scene recognition ability. The advantage of visual SLAM is the rich texture information it utilizes. For example, two billboards of the same size but different content can not be distinguished by the point cloud-based laser SLAM algorithm, but the vision can be easily distinguished. This brings unparalleled advantages in repositioning and scene classification. At the same time, visual information can be easily used to track and predict dynamic targets in the scene, such as pedestrians, vehicles, etc., which is crucial for the application in complex dynamic scenes.

The classic structure of visual SLAM system can be divided into five parts: camera sensor module, front-end module, back-end module, loop module and map building module. The camera sensor module is responsible for collecting image data, which can be a single camera or an array of cameras, to capture visual information in the environment. [4]The front-end module is responsible for tracking the image features between adjacent frames, achieving initial camera motion estimation and local map construction through feature matching, visual odometer and other technologies. The back-end module is responsible for numerical optimization of the data output by the front-end module, and further improves the accuracy of positioning and map through optimization algorithms (optimization in Figure 1, nonlinear optimization, etc.). The loop detection module is responsible for detecting the loop closure in the process of motion, eliminating the accumulated error by calculating the image similarity in large-scale environment, and improving the robustness and accuracy of the system. The mapping module is responsible for rebuilding the 3D map of the surrounding environment according to the data provided by the front-end and back-end modules, as well as marking and positioning the key points and road signs in the environment. Combining the functions of these modules, the visual SLAM system can realize autonomous positioning and mapping in unknown environments, and provide key position perception and environment understanding capabilities for automatic driving, robot navigation and other applications(figure1).

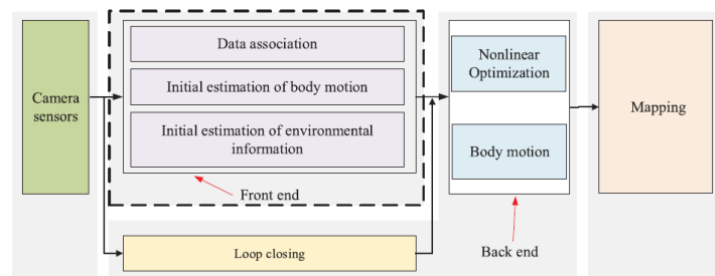

**Figure 1.** Visual SLAM system architecture diagram

1) Front end

The front end of visual SLAM is called visual odometer (VO). It is responsible for roughly estimating camera motion and feature direction based on information from adjacent frames. In order to obtain accurate attitude with fast response speed, an effective VO is required. Currently, front ends can be mainly divided into two categories: feature-based methods and direct methods (including semi-direct methods) (Zou et al., 2020). This section mainly reviews the feature-based approach to VO. Semi-direct and direct methods are discussed later.

VO systems based on feature points operate more stably and are less sensitive to light and dynamic targets. Feature extraction with high scale and good rotational invariance can greatly improve the reliability and stability of VO systems. In 1999, Lowe proposed the scale-invariant Feature Transform (SIFT) algorithm, which was improved and developed in 2004[5-6]. The whole algorithm is divided into three steps to complete the extraction and description of image feature points. (i) The scale space is constructed by Gaussian difference pyramid method, and the points of interest are identified by Gaussian differential functions. (ii) Determine the position and proportion of each candidate, and then identify the key points. (iii) Assign a pointing feature to a key point to obtain a descriptor.

2) Back end

The back-end receives the camera pose estimated by the front-end and optimizes the initial pose to obtain a globally consistent motion trajectory and environment map. Compared with the diverse algorithms on the front end, the current types of back-end algorithms can be mainly divided into two categories: filter-based methods (such as extended Kalman filter (EKF) Bailey et al., 2006) and optimization-based methods. They are described as follows:

A filter-based approach that mainly uses Bayesian principles to estimate the current state based on previous states and current observed data . Typical filter-based approaches include extended Kalman filters (EKF)  untracked Kalman filters (UKF)[7] , and particle filters (PF)  Taking the typical EKF-based SLAM method as an example, its application in small-scale environment is relatively successful. However, because the covariance matrix is stored and its storage capacity increases with the square of the amount of states, its application in large unknown scenarios is always limited.

The core idea of the optimization based method is to convert the back-end optimization algorithm into the form of a graph, which takes the subject pose and environmental characteristics at different moments as the vertices, and the constraint relationship between the vertices is represented by edges (Liang et al., 2013). After constructing the graph, an optimization-based algorithm is used to solve the pose of the target so that the state to be optimized on the vertex better satisfies the constraints on the corresponding edge. After the optimization algorithm is executed, the corresponding graphs are the target motion trajectory and the environment graph. At present, most mainstream visual SLAM systems use nonlinear optimization methods.

Therefore, in terms of actual data, visual SLAM systems typically use image data taken from a single or multiple cameras as input. These image data can be color images, depth images, or RGB-D images. By analyzing these image data, the visual SLAM system can extract image features, calculate camera motion, build environment maps, etc. At the same time, the visual SLAM system can also be fused with other sensor data, such as LiDAR data, inertial Measurement unit (IMU) data, etc., to improve the accuracy and robustness of positioning and mapping.

*2.4. The importance of automatic lane changes*

Automatic lane change is an important function of automatic driving system, and its design should consider the influence of urban road lights, obstacles and other traffic vehicles on lane change behavior.[8] In order to effectively formulate lane change rules for specific working conditions, the driver's lane change behavior can be simulated, and the influence of surrounding vehicles on lane change vehicles can be considered. Rule-based lane change behavior decision-making performs well in simple traffic scenarios, but has limitations in complex situations, such as failure to consider factors such as driver personality characteristics and road conditions, and bottlenecks in performance improvement.

In order to overcome these limitations, convolutional neural network is an effective method to extract information from sensor data and input it into decision network. The decision network generates speed and steering control instructions according to input information, and combines the security analysis method based on repulsive potential field to ensure the safety and humanization of automatic lane change. This approach allows for a more comprehensive consideration of the surrounding environment and vehicle dynamics, improving the performance and reliability of automated lane change systems for a safer and smarter autonomous driving experience.

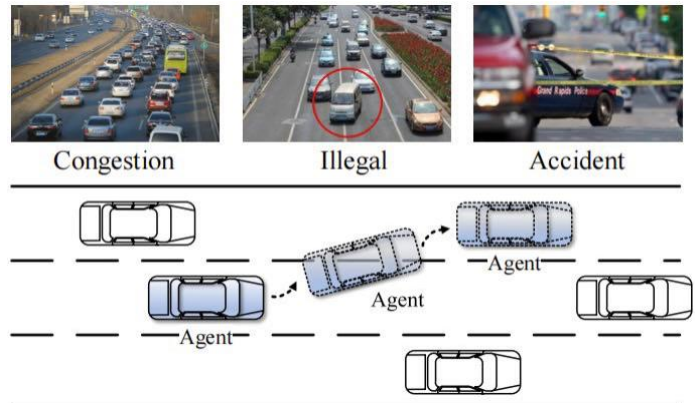
**Figure 2.** Auto lane change model diagram

Therefore, SLAM technology plays an important role in the automatic lane change of vehicles, and its key is to provide accurate environment perception and vehicle positioning, and provide necessary information support for the decision and control of automatic driving system(figure2). First of all, SLAM technology can build and update the environment map around the vehicle in real time, including road structure, obstacles, traffic lights, etc., which provides accurate scene cognition for automatic lane change. Secondly, SLAM technology can realize high-precision positioning of vehicle position, including the position of the vehicle lane and the relative position relationship with the surrounding vehicles, so as to ensure the accuracy and safety of lane change decision. In addition, SLAM technology can also monitor and predict the motion state of surrounding vehicles in order to adjust lane change strategies in time and avoid collisions with other vehicles.

For example, Waymo is an autonomous driving company that has applied SLAM technology, and its automatic lane change function relies on SLAM technology's perception of the surrounding environment and vehicle positioning. Waymo's system updates a map of the vehicle's surroundings in real time and uses SLAM technology to accurately estimate the vehicle's location and speed. During automatic lane changes, [7-9]Waymo's system analyzes the movement trajectory and speed of surrounding vehicles based on the information provided by SLAM technology to determine the best lane change timing and path, and ensures the safety of lane changes by maintaining a safe distance from surrounding vehicles. Through the application of SLAM technology, Waymo's automatic lane change function can achieve safe and efficient lane changes in complex traffic environments, improving the performance and reliability of the autonomous driving system

The above is an overview of the role and importance of SLAM technology in automatic lane change. Next, consider using Tesla's automatic lane change principle for experimental research to verify the effectiveness and reliability of SLAM technology in practical applications. Tesla's autopilot system uses advanced sensors and AI algorithms for environment perception and vehicle control, and its automatic lane change function is an important part of the autopilot system. By conducting experiments on the Tesla system, the role of SLAM technology in automatic lane change can be further verified, and its applicability and performance in different scenarios can be explored.

## 3. The application of SLAM technology in Tesla automatic lane change

Tesla recently announced a high-profile update to its Autopilot system, adding an automatic lane change feature. The launch of this feature means that Tesla cars can autonomously change lanes on highways without human intervention. [10]Specifically, this feature allows the vehicle to automatically overtake slow-moving vehicles in front of it, thereby improving the comfort and efficiency of driving. It is worth mentioning that the realization of this function involves the use of SLAM (Simultaneous Localization and Mapping) technology, through SLAM technology, Tesla cars can accurately perceive the surrounding environment, including vehicle location, road structure and traffic conditions, so as to make intelligent lane change decisions. At present, this feature has begun to push the upgrade through OTA (Over-The-Air) in the United States, bringing a more convenient and safe driving experience for Tesla owners. In this section, we will delve into the application of SLAM technology behind Tesla's automatic lane change feature and its importance in environmental perception.

### 3.1. Tesla Sensors detect the surrounding environment

Tesla vehicles are equipped with a variety of sensors, including radar, cameras, ultrasonic sensors, and LiDAR. These sensors are responsible for detecting vehicles, obstacles and road signs around the vehicle. There are a large number of LiDAR [11-13]sensors on the market today. Although all of these share the same light-based operating properties, manufacturers have offered different approaches to measurement and imaging systems, which translate into different performance metrics and overall costs. As shown in Figure 3, the LiDAR sensor uses an optical signal to measure distance, which is calculated based on the round-trip delay ($\tau$) between the signal emitted by the laser and the signal reflected by the target. Since the speed of light (c) is previously known, the distance to the target (R) can be calculated using an equation.

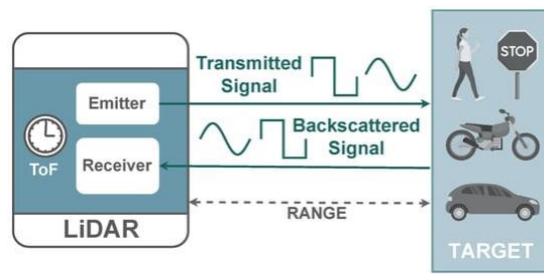

**Figure 3.** How LiDAR works

According to the principle shown in Figure 3, Tesla's autopilot system uses a variety of sensor technologies to sense the surrounding environment, including radar, cameras, ultrasonic sensors, etc. Among these sensor technologies, different methods may be used to measure the round-trip delay, known as time of flight (ToF), to obtain information about the distance between the target object and the vehicle.

A. Pulse signal technology: This technology sends a short pulse signal and measures the round-trip time of the signal to calculate the distance. This method is simple and easy to use, but easy to be affected by the signal-to-noise ratio, which limits the accuracy of measurement.

b. Amplitude-modulated continuous wave (AMCW) technology: The AMCW system does not send short pulses, but sends a modulated continuous wave signal. The sensor is synchronized with a series of integrating Windows and then calculates the distance based on the energy ratio present on each window. This method can obtain a better signal-to-noise ratio than the pulse signal technology, but its application scope is usually limited by wavelength, mainly for medium-range and short-range sensors.

c. Frequency-modulated continuous wave (FMCW) technology: FMCW systems use a continuous wave signal that increases in frequency over time, and then calculate the difference between the frequency of the returned signal and the local oscillator to derive distance information. Compared to the first two technologies, [14]FMCW technology provides greater robustness and can even detect the speed of moving targets.

Tesla's Autopilot system may utilize one or more of these sensor technologies to sense its surroundings and perform actions such as automatically changing lanes. The choice of these sensor technologies may take into account factors such as cost, performance, and scope of application to enable a safe and efficient autonomous driving experience.

### 3.2. Tesla LiDAR data collection

As one of the key sensors, LiDAR outputs point cloud data, which is an accurate 3D image of the vehicle's surroundings. This point cloud data plays a crucial role in autonomous driving tasks, such as object detection and classification, and collision avoidance.

Object detection and classification: Object detection and classification is an important task in LiDAR output point cloud data. First, the raw data needs to be transformed into a point cloud structure, and then point clustering or segmentation is performed to group the points according to common characteristics. This step identifies the set of points that may represent the object. Further processing, such as filtering or deleting redundant data, can then be done to reduce the amount of data to be processed later. In a fixed-location LiDAR application, the algorithm might first classify the point cloud into background and foreground, then cluster the foreground points and label a bounding box to represent a group of points that may represent an object.

**Object detection in moving vehicles:** Unlike LiDAR in fixed locations, LiDAR mounted on moving vehicles requires more robust and fast algorithms to process data because objects in the surrounding environment change at a higher frequency. [15]Traditional approaches may include the use of support vector machine (SVM) classifiers, as well as 2D or 3D representations based on machine learning techniques. These algorithms are designed to identify and classify objects in the point cloud to help vehicle systems better understand their surroundings.

Therefore, on Tesla vehicles, the point cloud data captured by LiDAR sensors is processed and analyzed by complex algorithms to identify various objects and road features in the surrounding environment, thus achieving functions such as object detection and classification. These functions are key components of autonomous driving systems that help vehicles drive intelligently and avoid collisions.

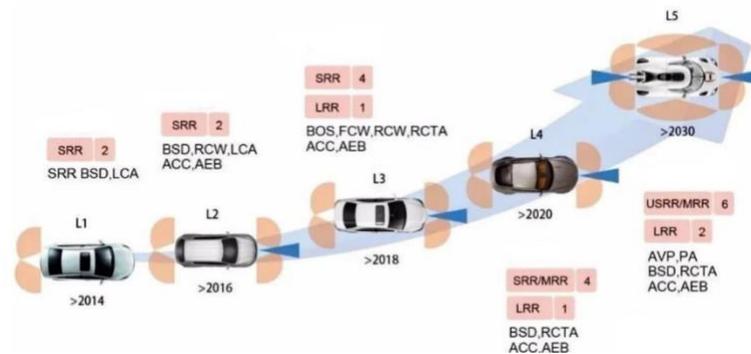

**Figure 4.** LiDAR sensors capture the surrounding environment

Let's say a Tesla vehicle is driving on a city road using autopilot. It is equipped with several sensors, including laser radar (LiDAR). The figure 4 vehicle is approaching an intersection. LiDAR sensors constantly scan the surrounding environment, and the captured point cloud data is sent to a computer system on the Tesla vehicle for processing and analysis.

*3.3. Case scenario*

Approaching an intersection, a LiDAR sensor detects a pedestrian at the intersection, as well as another vehicle coming from the side. By analyzing point cloud data in real time, Tesla vehicles' autopilot systems are able to accurately identify these objects and predict their behavior. For example, the system can predict whether a pedestrian intends to cross the road, as well as the speed and direction of oncoming cars.

**Decisions and controls:** Based on analysis of the surrounding environment, Tesla vehicles'[16] autopilot systems make decisions, such as adjusting the vehicle's speed and heading to ensure safe passage through intersections. If a potential collision risk is detected, the system may take emergency action, such as slowing down or performing evasive action.

This case demonstrates the use of LiDAR sensors in Tesla vehicles. By capturing and analyzing the point cloud data of the surrounding environment in real time, Tesla's autopilot system is able to accurately identify and classify various objects and make corresponding decisions to ensure the safe driving of the vehicle. Give a schematic that fits this case.

At the same time, LiDAR sensors are also widely used in SLAM (Simultaneous Localization and Mapping) technology. In SLAM, the LiDAR sensor is not only able to sense the structure of the surrounding environment, but also to realize the positioning and map construction of the vehicle in the unknown environment by measuring changes in its own position and attitude. By combining LIDAR-scanned point cloud data with the vehicle's motion trajectory, SLAM algorithms are able to update the vehicle's location information in real time and generate high-precision maps.

Therefore, the application of Lidar sensors in the field of autonomous driving is not limited to object detection and distance measurement, but also includes the support of SLAM technology, providing vehicles with more accurate and reliable positioning and navigation capabilities.

*3.4. Tesla path planning and control execution*

The vehicle's path planning and control execution is a complex system based on advanced sensor technology and intelligent algorithms. These sensors include [17]LiDAR, cameras, radar, and ultrasonic sensors that continuously scan and capture environmental data around the vehicle. When this data is processed and analyzed in real time, the system is able to accurately identify surrounding vehicles, pedestrians, road signs and obstacles, and assess road conditions.

Based on the analysis of sensor data, the autonomous driving system utilizes advanced path planning algorithms to determine the best driving path. These algorithms take into account the vehicle's current location, target location, road rules, traffic flow and other factors to ensure safe and efficient driving. For example, on a highway, the system might choose to follow the car in front to maintain a safe distance, or find the shortest alternative route in a traffic jam.

When the vehicle needs to change lanes, the automatic driving system will evaluate the traffic situation in the current lane in real time and plan the most appropriate lane change strategy. This may include waiting for the right distance and speed to move slowly into an adjacent lane, or using a gap to quickly change lanes. During this process, the system takes into account the speed, distance and direction of other vehicles, as well as the impact of the lane change action on surrounding vehicles, to ensure a safe and smooth lane change operation.

For example, when an autonomous vehicle is driving on a city road, if the vehicle in front suddenly brakes or a pedestrian suddenly crosses the road, the sensor will immediately detect and send an alert to the system. The system will analyze the surrounding environment and quickly re-plan the driving path to avoid collisions and ensure the safe passage of vehicles. If a lane change is needed to bypass a road obstacle, the system assesses the position and speed of surrounding vehicles based on traffic conditions and selects the best lane change timing and strategy to ensure safety and efficiency.

*3.5. Application conclusion*

In the Tesla automatic lane change function, SLAM (simultaneous localization and map construction) technology plays a key role. Tesla recently rolled out a major update to its Autopilot system, adding an automatic lane change feature. The launch of this feature means that Tesla cars can autonomously change lanes on highways without human intervention. Specifically, this feature allows the vehicle to automatically overtake the slow-moving vehicle ahead, thereby improving the comfort and efficiency of driving. It is worth mentioning that the realization of this function involves the use of SLAM (simultaneous positioning and map construction) technology, through SLAM technology, Tesla cars can accurately perceive the surrounding environment, including vehicle location, road structure and traffic conditions, so as to make intelligent lane change decisions. At present, the feature has begun to roll out upgrades through[18-19] OTA (over-the-air) in the United States, bringing a more convenient and safe driving experience for Tesla owners. By summarizing the application of SLAM in automatic lane change function and its importance in environment perception, we can further explore the conclusions of the third part.

Tesla's automated lane change feature is a complex system based on advanced sensor technology and intelligent algorithms. These sensors include LiDAR, cameras, radar, and ultrasonic sensors that continuously scan and capture data about the environment around the vehicle. When this data is processed and analyzed in real time, the system is able to accurately identify surrounding vehicles, pedestrians, road signs and obstacles, and assess road conditions.

SLAM technology plays a key role in Tesla's automatic lane change feature. Through sensors such as LiDAR, Tesla vehicles can not only perceive the structure of the surrounding environment, but also realize positioning and map construction in an unknown environment by measuring changes in their own position and attitude. By combining LIDAR-scanned point cloud data with the vehicle's motion trajectory, SLAM algorithms are able to update the vehicle's location information in real time and generate high-precision maps. [20]This allows Tesla vehicles to more accurately sense their surroundings, allowing them to make intelligent lane change decisions and ensure safe passage.

Therefore, Tesla's automatic lane change function not only relies on sensor technology and intelligent algorithms, but also relies on SLAM technology to provide a more accurate and reliable environment perception ability for the vehicle. This combination enables Tesla's Autopilot system to operate safely and efficiently in complex road environments, providing drivers with a more comfortable and convenient driving experience.

## 4. Conclusion

In terms of automatic lane change behavior prediction and environment perception based on SLAM technology, future development will focus on improving the intelligence and adaptability of the system. With the continuous advancement of artificial intelligence and sensor technology, automated lane change systems will more accurately identify obstacles and vehicles in complex road environments and be able to adjust the vehicle's trajectory in real time to ensure safe passage. A new generation of sensor technologies, such as deep learning-based cameras, millimeter-wave radars, and ultrasonic sensors, will provide richer and more comprehensive environment-aware data to the system, making automated lane changes smarter and more flexible. At the same time, SLAM technology will continue to play a key role in providing more reliable support for automated lane change decisions through more accurate environmental maps and vehicle positioning information. Future research will also explore how to further optimize SLAM algorithms to adapt to a wider range of road conditions and complex traffic situations, enabling reliable and efficient driving of autonomous vehicles in a variety of scenarios.

In addition, with the continuous maturity and popularization of automatic lane change technology, the requirements for system safety and reliability will be increasingly high. Therefore, future research will also focus on how to further improve the predictive accuracy and environmental awareness of automated lane change systems through methods such as reinforcement learning and deep learning. By integrating AI technology and sensor data, automated lane-changing systems will be able to more accurately predict the behavior of other vehicles and take appropriate measures when necessary to ensure driving safety. With continuous improvements to SLAM technology and automated lane change systems, we can look forward to a future where autonomous vehicles are safer and more efficient on the road, providing drivers with a more comfortable and convenient driving experience.